\pdfoutput=1

\documentclass[11pt]{article}

\usepackage{acl}

\usepackage{times}
\usepackage{latexsym}

\usepackage[T1]{fontenc}

\usepackage[utf8]{inputenc}

\usepackage{microtype}

\usepackage{inconsolata}

\usepackage{colortbl}
\usepackage{xcolor}
\definecolor{Gray}{gray}{0.85}  
\usepackage{graphicx}
\usepackage{newfloat}
\usepackage{listings}
\usepackage{booktabs} 
\usepackage{array}
\usepackage{svg}
\usepackage{float}
\usepackage{enumitem}
\usepackage{tabularx} 
\usepackage{amsmath}
\usepackage{appendix}

\newcolumntype{L}{>{\raggedright\arraybackslash}X} 
\newcolumntype{C}{>{\centering\arraybackslash}m{1.5cm}}  

\title{DataFrame QA: A Universal LLM Framework on DataFrame Question Answering Without Data Exposure}

\author{Junyi Ye\textsuperscript{1}, Mengnan Du\textsuperscript{1}, Guiling Wang\textsuperscript{1}\\
  \textsuperscript{1}New Jersey Institute of Technology\\
  \small\texttt{\{jy394,mengnan.du,gwang\}@njit.edu}
}

\begin{document}
\maketitle
\begin{abstract}
This paper introduces DataFrame question answering (QA), a novel task that utilizes large language models (LLMs) to generate Pandas queries for information retrieval and data analysis on dataframes, emphasizing safe and non-revealing data handling. Our method, which solely relies on dataframe column names, not only ensures data privacy but also significantly reduces the context window in the prompt, streamlining information processing and addressing major challenges in LLM-based data analysis. We propose DataFrame QA as a comprehensive framework that includes safe Pandas query generation and code execution. Various LLMs, notably GPT-4, are evaluated using the pass@1 metric on the renowned WikiSQL and our newly developed `UCI-DataFrameQA', tailored for complex data analysis queries. Our findings indicate that GPT-4 achieves pass@1 rates of 86\% on WikiSQL and 97\% on UCI-DataFrameQA, underscoring its capability in securely retrieving and aggregating dataframe values and conducting sophisticated data analyses. This approach, deployable in a zero-shot manner without prior training or adjustments, proves to be highly adaptable and secure for diverse applications.
\end{abstract}

\section{Introduction}
In the era of large language models (LLMs), table question answering (QA) with LLMs typically involves embedding the entire table into the prompt, along with the user's question and instructions \cite{li2023table,chen2022large}. This method is highly effective for querying small and simple tables or dataframes. However, since tables are inherently two-dimensional structures, they can quickly increase the size of the prompt with the addition of rows. This becomes particularly challenging with large dataframes such as those related to weather, traffic, and product sales, which can easily exceed the 4K or 8K content window limit of most models.

\begin{figure}[ht]
\centering
\includegraphics[width=\linewidth]{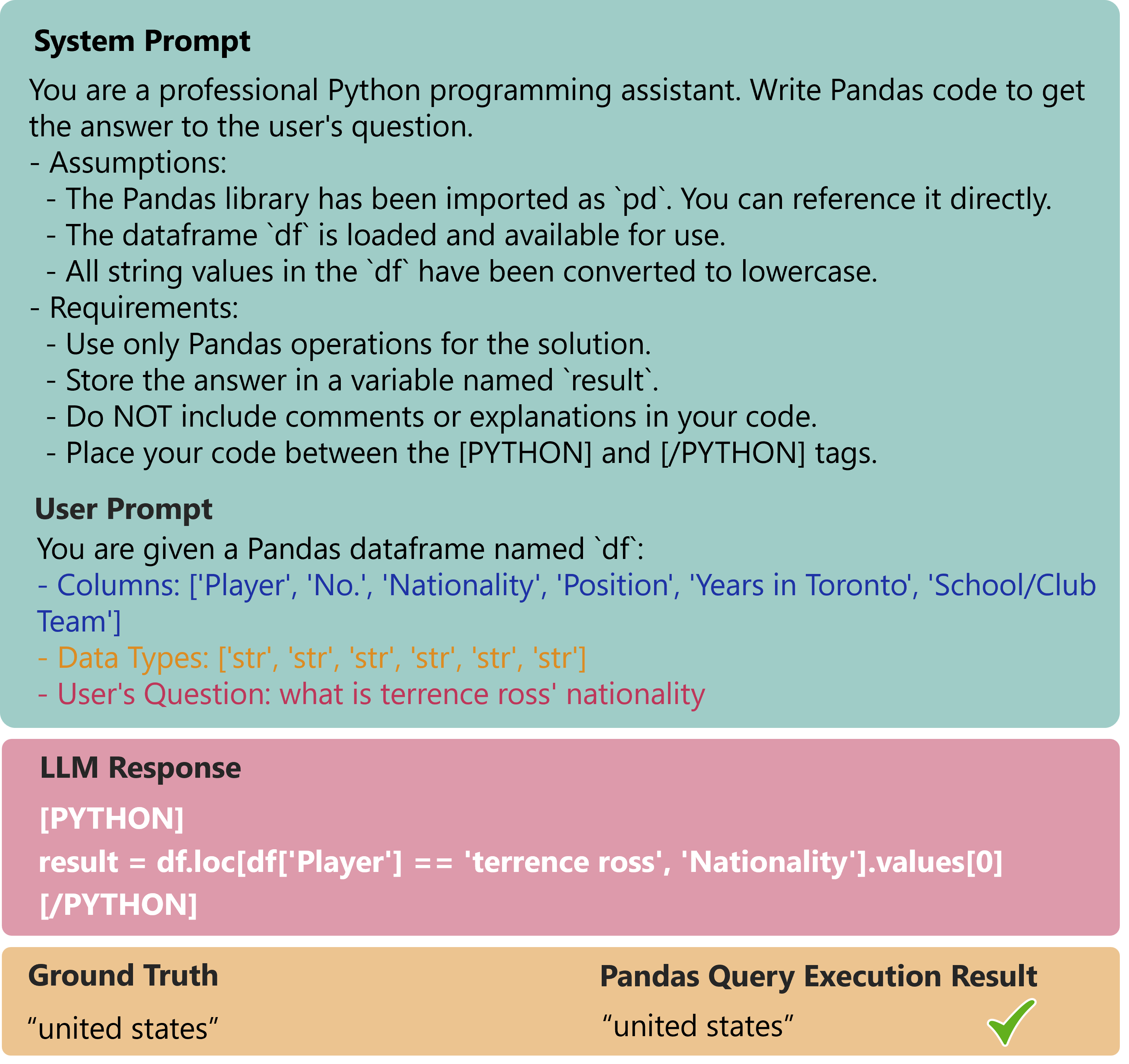}
\caption{An example where a LLM can generate a correct Pandas query to answer user question using only the table header and column data types, without accessing the table values. Typically, the total number of tokens for DataFrame QA tasks, including both the prompt and the model output, stays below 250 tokens.}
\label{fig:prompt_code_generation}
\end{figure}

Unlike lengthy text content that can be efficiently managed using techniques such as Retrieval-Augmented Generation (RAG) \cite{lewis2020retrieval}, which allows the summarization of each chunk and integrates searching to generate a final answer (thus efficiently processing a large volume of text), large tables present a different challenge. Tables, due to their two-dimensional structure and data density, do not lend themselves to this kind of summarization and retrieval-based processing. This difference highlights the unique challenge of using LLMs to efficiently manage large table data, as opposed to handling extensive text content.

Despite many models (e.g., GPT-4 and Claude-2.1) now expanding their context window sizes to 16k, 32k, or more, several key challenges remain. First, the computational cost for processing tokens is high. For example, it costs \$0.01 per 1K tokens using the latest GPT-4 Turbo model API. Second, embedding full datasets risks potential data leakage. In addition, recent studies indicate the problem of `Lost in the Middle,' where long prompts can decrease the model performance \cite{liu2023lost}. LLMs also struggle with mathematics, which becomes problematic when the query involves calculations \cite{ouyang2022training}. Moreover, conventional querying often includes superfluous data beyond what is necessary to answer the question. 
These pose challenges in maintaining accuracy, efficiently managing computational resources and protecting sensitive information when applying LLMs to table-based QA. 

To address these challenges, we propose a new task and framework called DataFrame QA. This framework aims to enhance the efficiency and security of querying the dataframe by using LLM to generate Pandas queries. It employs a method that utilizes only table column names and data types, effectively reducing data leakage risks and minimizing the need for extensive context windows. We modify the WikiSQL dataset and create a new dataset, UCI-DataFrameQA for our task, conducting evaluations in a zero-shot manner using Llama2 \cite{touvron2023llama}, CodeLlama \cite{roziere2023code}, GPT-3.5, and GPT-4~\cite{openai2023gpt4}. 
Furthermore, our research involves a thorough analysis of the causes of errors and the inherent challenges associated with the DataFrame QA task, along with potential solutions. This investigation provides insights for future dataset expansions and improvements and for enhancing model performance.

\section{Task Setups and Challenges}
The DataFrame QA task introduces a novel challenge in Natural Language Processing (NLP), focusing on employing LLMs to generate Pandas queries. This highlights the versatility of Pandas, which offers a wide array of data analysis operations, ranging from simple data retrieval to intricate statistical analyses.

\subsection{DataFrame QA Task}
This task marks a departure from traditional data processing, which typically involves analyzing the relationship between questions and table contents. The DataFrame QA task, instead, concentrates on the analysis of dataframe structures and data types, deliberately omitting the scrutiny of actual data values. It evaluates LLMs' ability to understand dataframe headers and column metadata, and to convert this understanding into valid Pandas queries through natural language questions.

Moreover, the task utilizes the versatility of prompts to include additional information, such as dataset descriptions and specific constraints. This enhances contextual comprehension beyond what traditional fixed-input models offer. This approach facilitates a more dynamic and informed model interaction, greatly expanding the potential of dataframe analysis in NLP.

The input to the DataFrame QA system is defined as a tuple \((S, H, C, Q)\), where \(S\) represents supplementary information, like assumptions and Python library constraints, \(H = \{h_1, h_2, ..., h_n\}\) denotes the dataframe headers, and \(C = \{c_1, c_2, ..., c_n\}\) covers additional column information, such as data types and descriptions. \(Q\) is the natural language query. The system's output, \(P'\), is a Pandas query generated in response to \(Q\). The ground truth query, \(P\), is the correct Pandas query that yields the answer to \(Q\). Hence, the predicted output is a function of the input:

\begin{align}
&P' = f(S, H, C, Q) \\
&A' = execute(P', df) \\
&A = execute(P, df)
\end{align}

\(A'\) and \(A\) are the results obtained by executing \(Q'\) and \(Q\) on the dataframe \(df\) in a safe sandbox, respectively. The effectiveness of the DataFrame QA system  is measured by how closely \(A'\) approximates the ground truth answer \(A\).

\begin{figure*}[ht]
\centering
\includegraphics[width=0.8\linewidth]{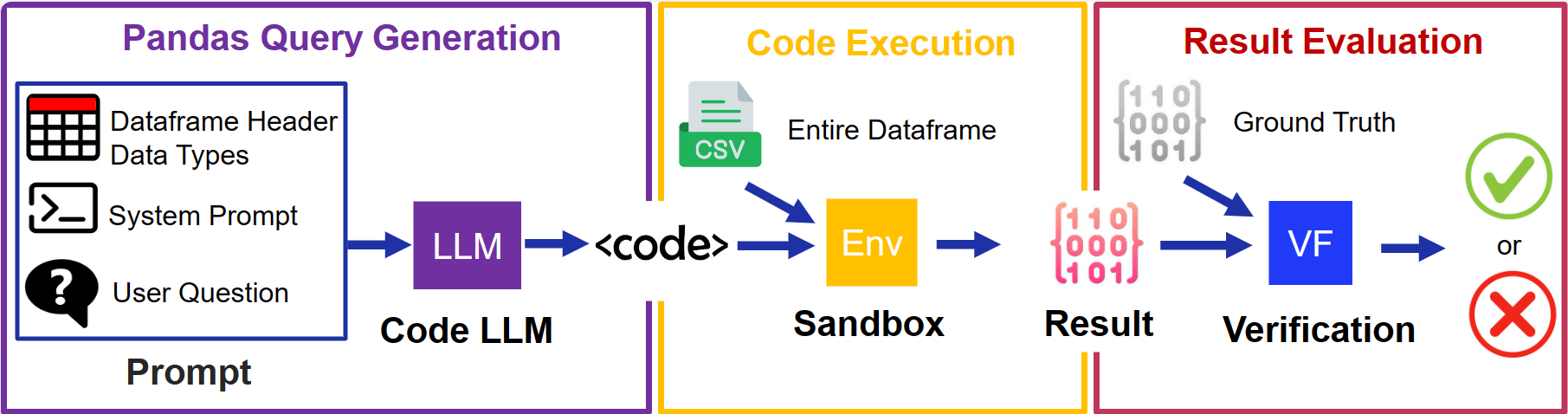}
\caption{Framework of DataFrame QA.}
\label{fig:framework}
\end{figure*}

\subsection{Challenges}
Key challenges of DataFrame QA include:

\begin{itemize}[label={},leftmargin=0pt]
\item \textbf{Interpreting User Questions:} Understanding how questions relate to the dataframe's structure and column types, requiring domain-specific knowledge and handling ambiguities.
\item \textbf{Formulating Pandas Queries:} Proficiency in creating accurate Pandas queries that meet the technical and logical requirements of the task.
\item \textbf{Following Instructions:} Adhering to given guidelines or constraints, ensuring that responses are technically correct and contextually suitable.
\end{itemize}
These challenges highlight the intricacies involved in the Dataframe QA task, underscoring the need for LLMs that are not only proficient in natural language processing but also capable of understanding and manipulating complex data structures.

\section{Methodology}
Figure \ref{fig:framework} outlines our DataFrame QA framework, structured in three stages: Pandas Query Generation, Code Execution, and Result Evaluation.

\subsection{Pandas Query Generation}
In this initial phase, a LLM processes the prompt, comprising the dataframe header, column data types, system prompt (i.e. assumption and requirements), and user question, to generate a Pandas query. This design ensures data privacy by avoiding exposure to table values and leverages column data types to inform query selection.

\subsection{Code Execution}
The generated query is executed within a controlled virtual environment, protecting against unauthorized operations. This environment is restricted to essential libraries (Pandas, NumPy, and Math), thereby enhancing security. The execution results, stored as Python objects, offer flexibility for further processing. For example, large table results can be provided as downloadable content, while Matplotlib plot objects are displayed directly.

\subsection{Result Evaluation}
We compare the results of executed queries with ground truth answers, encountering challenges due to the diversity of data types involved. Our methodology standardizes results across numeric, string, and list/ndarray types to facilitate an accurate comparison. Note that Pandas queries often return series or dataframe objects, rather than direct answers to user questions, mirroring the characteristics of coding datasets commonly used in LLM training. To address this, we employ a relaxed evaluation criterion, considering the contents of series or dataframes correct if they include the answer. For pairs where there is a mismatch, we perform a manual comparison to ensure the accuracy and relevance of the results.


\section{Experimental Settings}

\subsection{Dataset}
To rigorously assess the proficiency of LLMs in generating Pandas queries for two distinct types of tasks, we have adapted the WikiSQL and UCI datasets to align with our research objectives.

\begin{table*}[ht]
\centering
\scriptsize 
\begin{tabularx}{\textwidth}{p{0.28\textwidth}X>{\raggedright\arraybackslash}p{1.4cm}}
\toprule
\textbf{User Question} & \textbf{Pandas Query} & \textbf{Types} \\
\midrule
which province is bay of islands in? & \texttt{result = df.loc[df[`Electorate']==`bay of islands', `Province'].iloc[0]} & Retrieval \\
\rowcolor{Gray}
how many combined days did go shiozaki have? & \texttt{result = df.loc[df[`Wrestler']==`go shiozaki', `Combined days'].values[0]} & Aggregation \\
how does the average shell weight vary across different numbers of rings? & \texttt{result = df.groupby(`Rings')[`Shell\_weight'].mean()} & Data Analysis \\
\rowcolor{Gray}
can you create a new column `volume' as a product of length, diameter, and height, then find the average volume for each sex? & \begin{tabular}[t]{@{}l@{}}\texttt{df[`Volume'] = df[`Length'] * df[`Diameter'] * df[`Height'];} \\ \texttt{result = df.groupby(`Sex')[`Volume'].mean()}\end{tabular} & Data Analysis \\
\bottomrule
\end{tabularx}
\caption{Examples of Sample Questions and Corresponding Pandas Queries Categorized by Complexity Level. Retrieval/Aggregation queries can be resolved using single-step, SQL-like queries, whereas Data Analysis questions necessitate multi-step or complex Pandas operations.}
\label{table:queries}
\end{table*}

\textbf{Simple Query Dataset - WikiSQL}:
WikiSQL \cite{zhong2017seq2sql}, a benchmark in Text-to-SQL research, provides a test set comprising 15,878 table-question pairs, designed to evaluate natural language interfaces with relational databases. We transformed these tables into dataframes, ensuring datatype consistency for each column. Additionally, we utilized the results of the SQL queries executed on these tables as the ground truth for our DataFrame QA framework.

A notable challenge in WikiSQL is the frequent lowercasing of entities in user questions, which can lead to ambiguities when formulating Pandas queries. To mitigate this issue, we standardized all user questions and dataframe strings to lowercase. Additionally, we explicitly instructed in the prompt that all strings within the dataframe are lowercased. This approach improves clarity and uniformity in LLM query processing, ensuring consistent interpretation and handling of string data.

The WikiSQL dataset predominantly features straightforward information retrieval questions (71\%), solvable with single-step operations similar to basic SQL queries. The remaining 29\% focus on aggregation tasks, including 12\% MIN/MAX, 9\% COUNT, and 8\% AVG/SUM. These represent simpler and more direct query scenarios, which require basic dataframe operations.

\textbf{Complex Dataset - UCI-DataFrameQA}:
To develop a DataFrame QA dataset reflective of real-world scenarios, we adopted a comprehensive approach. We sourced diverse dataframes from the UCI dataset \cite{ucidataset}, spanning various domains such as animals, automobiles, and medical fields, to simulate different societal contexts. Our methodology was designed to represent three real-life data interaction roles: \textbf{1) Data Scientists}, who delve into detailed data analysis queries for patterns, trends, and statistical insights; \textbf{2) General Users}, such as patients in medical datasets or customers in automobile datasets, seeking practical, consumer-oriented aspects of the data; \textbf{3) Data Owners}, like hospitals or companies, focusing on extracting business-oriented insights.

Utilizing GPT-4, we generated questions mirroring typical inquiries and challenges these roles face in real-life scenarios, thereby creating a comprehensive DataFrame QA dataset.

For each of the 11 dataframes from the UCI dataset, GPT-4 generated 60 questions (20 per role), each with a corresponding Pandas query. The appendix details the prompts and provides examples of the question/Pandas query generation.

Following a meticulous manual review, we compiled a final set of 547 question/Pandas query pairs. This curation involved eliminating pairs with inaccurate matches, those requiring external libraries, or questions unsolvable with just the provided table headers. This rigorous selection process ensures the dataset comprises realistic and executable DataFrame QA scenarios.

\begin{table}[h]
\centering
\small 
\begin{tabular}{lcc}
\toprule
Role & Retrieval/Aggregation & Data Analysis \\
\midrule
Data Scientist & 9 (5\%) & 175 (95\%) \\
General User & 69 (40\%) & 105 (60\%) \\
Data Owner & 42 (22\%) & 147 (78\%) \\
\bottomrule
\end{tabular}
\caption{Distribution of Generated Question Types on UCI Dataset Across Different Roles.}
\label{tab:query_distribution}
\end{table}

Table \ref{tab:query_distribution} presents the distribution of generated question types within the UCI dataset, categorized by different user roles. Analysis of the dataset indicates that 22\% of the questions are focused on basic retrieval/aggregation tasks, while a predominant 78\% involves more advanced operations, such as grouping, correlation analysis, and sorting. Specifically, the Data Scientist role primarily concerns complex data analysis queries, whereas General Users tend to concentrate on more straightforward retrieval/aggregation questions. Questions from Data Owners exhibit a range of complexity, bridging the two extremes. The dataset is a pivotal testbed for assessing LLMs' adeptness in handling advanced queries, showing their capacity to execute complex, multi-step data analysis.

This distinction in question complexity across two datasets provides an opportunity to evaluate LLMs' capabilities over a wide spectrum of query complexities, ranging from simple data retrieval to sophisticated data manipulation tasks.

\subsection{Baselines}
Our experimental baselines include:

\begin{itemize}[label={},leftmargin=0pt]
\item \textbf{Llama2.} This advanced iteration of the Llama language model series offers configurations ranging from 7B to 70B parameters \cite{touvron2023llama}. With training on 2 trillion tokens, it is equipped with an expanded 4K token context window, enhancing its applicability across diverse NLP tasks.

\item \textbf{CodeLlama.} A specialized variant of Llama2, CodeLlama is tailored for coding-related tasks \cite{roziere2023code}. It demonstrates superior performance in coding benchmarks, benefitting from a 16K token window. We utilized its instruction models at 7B, 13B, and 34B parameter sizes, focusing on their ability to generate and interpret code.

\item \textbf{GPT-3.5, GPT-4.} These models are benchmarks in the LLM domain, showcasing exceptional performance across coding, general NLP tasks, and conversational capabilities. Their ongoing updates reinforce their status as leaders in AI-driven solutions. The model versions used in our experiments are gpt-3.5-turbo-0613 and gpt-4-0613.
\end{itemize}

\subsection{Implementation Details}
In our methodology, a consistent greedy decoding strategy was applied across all LLMs. For the deployment of Llama2 and CodeLlama, the official checkpoints available through the HuggingFace were used. The models were executed on NVIDIA A100 GPUs, with the number of units ranging from one to four, depending on the model size.

\subsection{Evaluation Metric}
The pass@1 score \cite{chen2021evaluating} is a crucial metric in evaluating the performance of LLM in the context of code generation tasks. It measures the accuracy of the LLMs in providing the correct answer on their first attempt without any additional iterations or refinements. This score is especially important in scenarios where immediate and precise responses are required, reflecting the model's ability to accurately interpret and respond to complex queries in a single pass. 

\section{Experimental Analysis}
This study delves into the DataFrame QA task, focusing on the capabilities and challenges of cutting-edge LLMs. Our investigation is centered around three pivotal research questions: \textbf{Q1. Are current state-of-the-art LLMs capable of effectively handling the DataFrame QA task? Q2. What factors influence the performance differences among various LLMs? Q3. What are the inherent challenges and potential solutions associated with DataFrame QA tasks?}

\begin{figure}[ht]
\centering
\includegraphics[width=\linewidth]{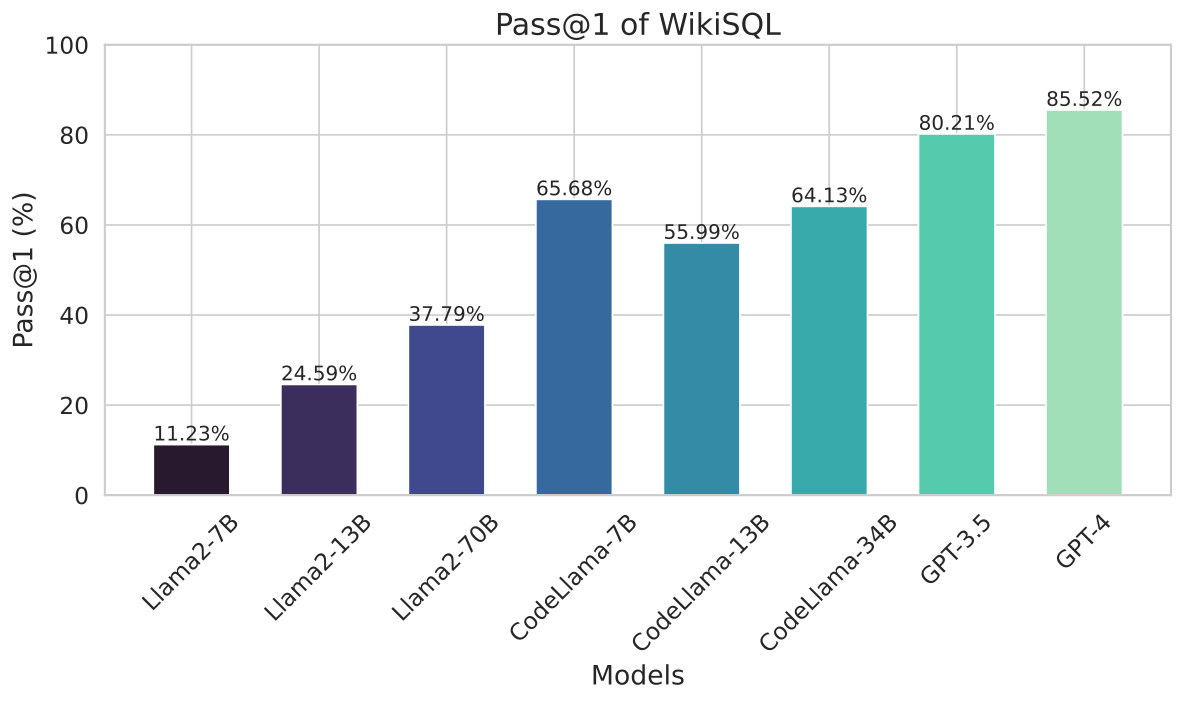}
\caption{Performance of LLMs on WikiSQL.}
\label{fig:pass@1_wikisql}
\end{figure}

\begin{figure}[ht]
\centering
\includegraphics[width=\linewidth]{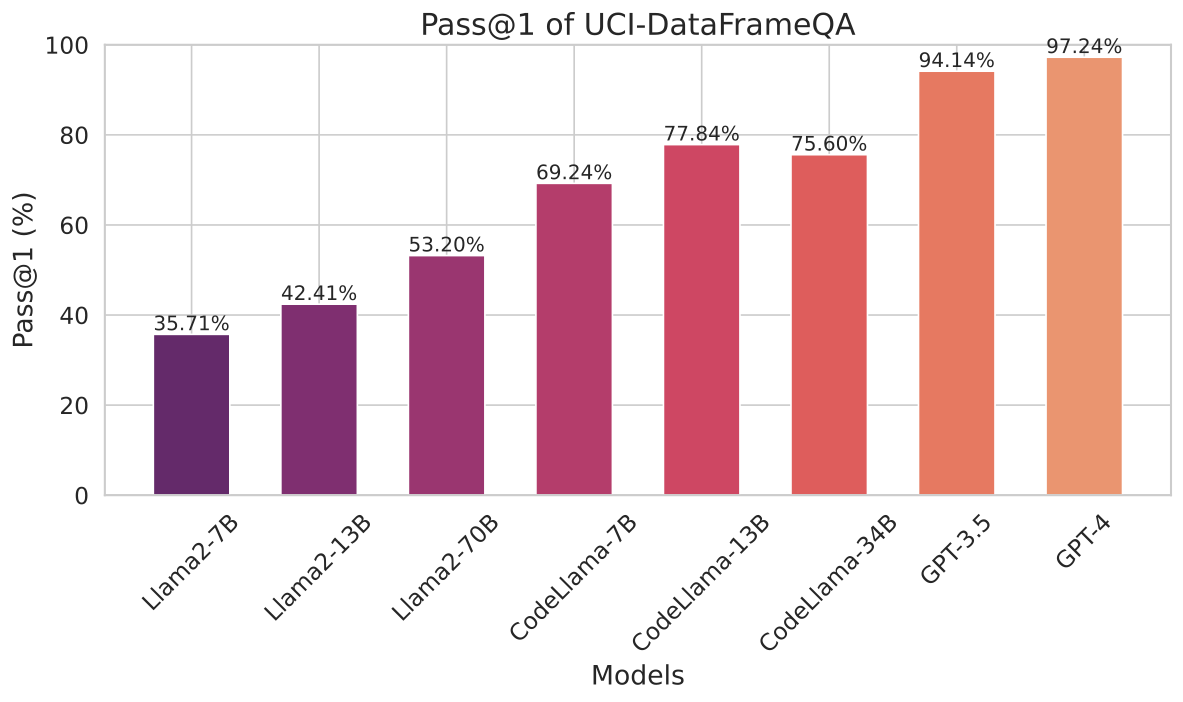}
\caption{Performance of LLMs on UCI-DataFrameQA.}
\label{fig:pass@1_uci}
\end{figure}

\begin{itemize}[label={},leftmargin=0pt]
\item \textbf{Q1:  Efficacy of Leading LLMs in DataFrame QA Tasks}
Our study assesses various LLMs' first-attempt accuracy (pass@1), with findings illustrated in Figures \ref{fig:pass@1_wikisql} and \ref{fig:pass@1_uci}. Notably, GPT-4 exhibits high pass@1 accuracies—85.5\% on WikiSQL and 97.2\% on UCI-DataFrameQA, reflecting its adeptness in processing a wide array of queries, from simple data retrieval to complex analysis, without direct access to table data.

Analysis of GPT-4 on WikiSQL's test set reveals that 11.6\% of queries pose challenges due to mismatched pairs (5\%), ambiguity (5\%), and quotation mark issues (1.6\%), affecting execution. Addressing these issues, GPT-4's pass@1 accuracy could rise to 96.7\%, aligning with its performance on UCI-DataFrameQA.

\noindent \textbf{Scaling Laws in LLM Performance: }
We observed clear performance stratification among models: Llama2 and CodeLlama, with the latter surpassed by the GPT series. GPT and Llama2 models adhere to scaling laws (\cite{kaplan2020scaling}), showing gains with increasing size. In contrast, CodeLlama models deviate from these laws, warranting further exploration in subsequent sections.

\noindent \textbf{Comparison with Text-to-SQL Models:}
GPT-4, in comparison to specialized Text-to-SQL models such as TAPEX \cite{liu2021tapex} with an execution accuracy of 89.5\% and SeaD+EG\textsubscript{\text{CS}} \cite{xu2021sead} at 92.7\%, exhibits slightly lower performance. This discrepancy is attributed to two main factors:
\textbf{Task Complexity:} Generating Pandas queries is inherently more complex than structured SQL queries, given Pandas' wider operation range.
\textbf{Zero-Shot Learning Approach:} Unlike TAPEX and SeaD+EG\textsubscript{\text{CS}}, which are specialized for Text-to-SQL, GPT-4's zero-shot application, without specific fine-tuning, impacts its efficiency in DataFrame QA tasks.

\begin{figure}[ht]
\centering
\includegraphics[width=\linewidth]{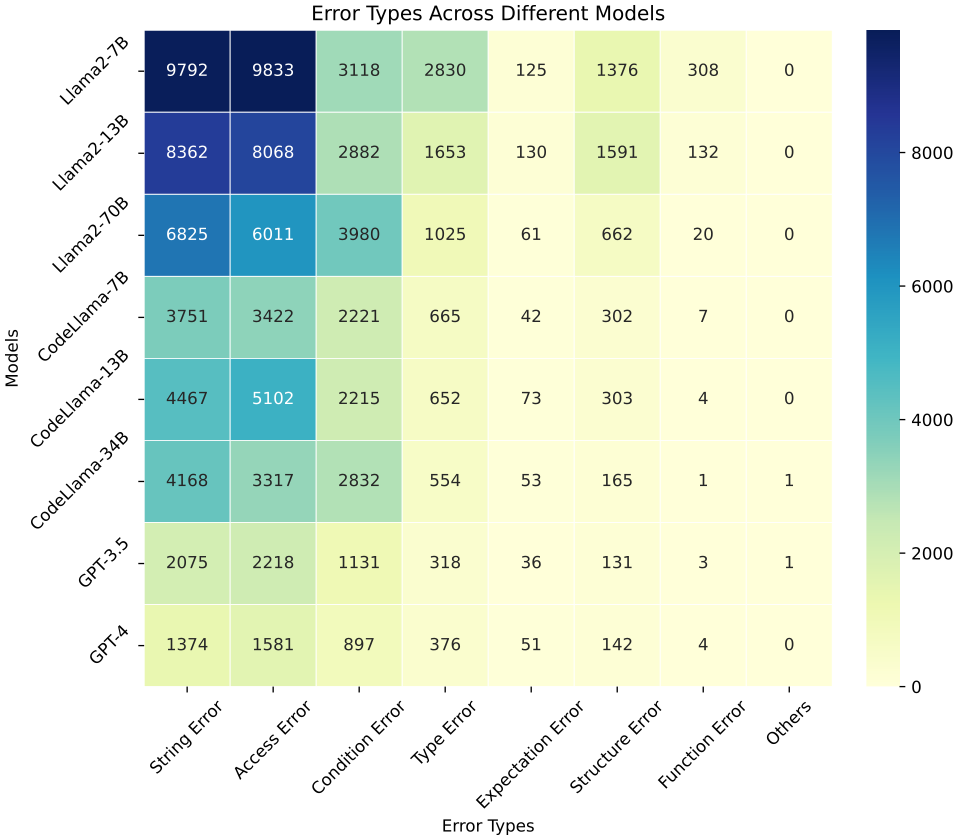}
\caption{Distribution of error types among different LLMs on WikiSQL. Definitions of error types and prompts for error classification are put in Appendix.}
\label{fig:heatmap}
\end{figure}

\item \textbf{Q2. Performance Variation Determinants}

In delving into what drives performance differences in LLMs, we scrutinize failure cases on WikiSQL using GPT-3.5. Starting with 100 error samples, we categorized these using GPT-4 and further confirmed through manual verification, identifying eight distinct error types detailed in the Appendix. We then used GPT-3.5 to classify all error classes across model input, sample rows, generated queries, execution output, and expected results, where incorrect queries often span multiple error categories.

\textbf{Error Distribution Across LLMs:}
Figure \ref{fig:heatmap} presents a heatmap of error class distribution across LLMs. Predominant errors include String Matching and Comparison, Data Access and Bounds, and Query Condition and Value. These stem from issues like misidentified column names, case sensitivity, and misinterpretations of user questions and instructions, often due to a mismatch between instructions and their execution.



\textbf{Bias Between Instructions and Training Data:}

In our analysis of CodeLlama models, we identified a notable bias reflecting discrepancies between instruction adherence and training data distribution, particularly evident in handling string queries. The 13B model registered 2368 instances, and the 34B model recorded 1175 instances of errors related to this bias, in stark contrast to the mere 175 instances encountered by the 7B model. These errors persisted even with explicit instructions to treat all strings as lowercase, underscoring a variance in the models' interpretation and compliance with these directives, deviating from the expected scaling law behavior in such tasks.

For instance, \texttt{df[df[`Player']==`Terrence Ross'][`Nationality'].values[0]} incorrectly queries the capitalized \texttt{`Terrence Ross'}, contrary to the directive of all strings in the dataframe being lowercase. This example highlights the bias issue: when querying dataframe values, especially with proper nouns like names and places, the model tends to capitalize them, showing a preference influenced by its training data over the provided instructions. This tendency exemplifies the broader challenge of aligning model behavior with specific directives, particularly when they counter the model's training data distribution. 

These results reveal that DataFrame QA tasks test not only the coding abilities of LLMs but also their ability to adhere to instructions in the face of conflicts with training data distribution, along with their comprehensive skills in understanding user queries and dataframe headers.

\begin{table*}[ht]
\tiny
\centering 
\caption{Typical Failure Cases in DataFrame QA Task.} 
\begin{tabular}{>{\raggedright\arraybackslash}p{1.5cm}p{2.5cm}p{5cm}p{5cm}}
\toprule
\textbf{Error Type} & \textbf{User Question} & \textbf{Generated Pandas Query} & \textbf{Expected Pandas Query}\\
\midrule

Value Retrieval Error & which province is grey and bell electorate in &  \begin{tabular}[t]{@{}l@{}}\texttt{result = df[(df[`Electorate']==\textbf{`grey'})|} \\ \texttt{(df[`Electorate']==\textbf{`bell'})][`Province'].unique()}\end{tabular} & \begin{tabular}[t]{@{}l@{}}\texttt{result=df[(df[`Electorate']==\textbf{`grey and bell'})]} \\ \texttt{[`Province'].unique()}\end{tabular} \\ 

\rowcolor{Gray}
Column Reference Error & what is the variance in resting blood pressure (trestbps) among different heart disease severity levels? &  \texttt{df.groupby(\textbf{`ca'})[`trestbps'].var()} & \texttt{df.groupby(\textbf{`num'})[`trestbps'].var()} \\ 

Instruction Misalignment & What are the mean and median lengths of abalone by each sex category? & \begin{tabular}[t]{@{}l@{}}\texttt{\textbf{import pandas as pd}} \\ \texttt{\textbf{\# Group the dataframe by sex}} \\ \texttt{...} \end{tabular} & \begin{tabular}[t]{@{}l@{}}\texttt{result = df.groupby(`Sex')[`Length'].agg(} \\ \texttt{[`mean', `median'])}\end{tabular} \\ 

\rowcolor{Gray}
Aggregation Error & what is the total amount of allied-unrelated where the component is human capital? & \begin{tabular}[t]{@{}l@{}}\texttt{result = df[df[`Component']==`human capital']} \\ \texttt{[`Allied-Unrelated'].\textbf{sum()}}\end{tabular} & \begin{tabular}[t]{@{}l@{}}\texttt{result = df[df[`Component']==`human capital']} \\ \texttt{[`Allied-Unrelated'].\textbf{count()}}\end{tabular} \\ 

Function-Column Ambiguity Error & what is the average score when the swimsuit score is 8.503 & \begin{tabular}[t]{@{}l@{}}\texttt{result = df[df[`Swimsuit']==`8.503']} \\ \texttt{[`Average'].\textbf{mean()}}\end{tabular} & \texttt{result = df[df[`Swimsuit']==`8.503'][`Average']} \\ 

\rowcolor{Gray}
Insufficient Column Data/Format Information & which team played on december 5? & \begin{tabular}[t]{@{}l@{}}\texttt{df[`Date'] = pd.\textbf{to\_datetime}(df[`Date'])} \\ \texttt{result = df[df[`Date']==`December 5'][`Team']}\end{tabular} & \texttt{result = df[df[`Date'] == `december 5'][`Team']} \\ 

Coding Syntax Error & how does the average shell weight vary across different numbers of rings? & \begin{tabular}[t]{@{}l@{}}\texttt{result = df.groupby(`Rings').\textbf{mean()}} \\ \texttt{[`Shell\_weight']}\end{tabular} & \begin{tabular}[t]{@{}l@{}}\texttt{result = df.groupby(`Rings')[`Shell\_weight']} \\ \texttt{.\textbf{mean()}}\end{tabular} \\ 

\rowcolor{Gray}
Hallucination Error & i'm interested in knowing the most common age of abalone. can you find that for me? & \texttt{result = df[\textbf{`Age'}].mode()}  & \texttt{result = df[\textbf{`Rings'}].mode()]} \\ 

\bottomrule
\label{tab:failure cases}
\end{tabular}
\end{table*}

\item \textbf{Q3. Inherent Challenges and Solutions in DataFrame QA Tasks}

In Table \ref{tab:failure cases}, we outline typical failure cases in DataFrame QA tasks, categorizing a spectrum of error types commonly encountered by LLMs. These include Value Retrieval, Column Reference, Instruction Misalignment, Aggregation, Function-Column Ambiguity, Insufficient Column Data/Format Information, Coding Syntax, and Hallucination Errors, each posing distinct challenges. Detailed examples and solutions for varied failure types are comprehensively provided in the Appendix.

\begin{itemize}[label={},leftmargin=0pt]
\item \textbf{Value Retrieval Error:}
This error occurs when Pandas queries fetch incorrect values due to ambiguous user questions. For instance, a query for \texttt{`grey and bell'} might be misinterpreted as separate searches for \texttt{`grey'} and \texttt{`bell'}. Also, LLMs might omit crucial characters like Roman numeral \texttt{`I'} or hyphens, mistaking them for textual errors.

\textit{Solution:} Utilizing quotation marks around specific query terms can significantly reduce these errors, clarifying the intended search as a singular entity.

\item \textbf{Column Reference Error:} 
Errors occur when queries incorrectly target columns, often due to ambiguous names. This common challenge arises especially when LLMs lack domain-specific knowledge, leading to column misidentification.

\textit{Solution:} Providing clear column descriptions in prompts, such as specifying the roles of \texttt{`ca'} and \texttt{`num'} columns, can greatly reduce these errors.

\item \textbf{Instruction Misalignment Error:}
This error occurs when LLMs deviate from given instructions, often seen in superfluous import statements and comments, typical in training datasets but unnecessary for DataFrame QA tasks. It also includes issues like case sensitivity errors, where LLMs incorrectly handle capitalized names or terms.

\textit{Solution:} Precise directives in prompts, such as \texttt{`Do not import Pandas library'} and specific case sensitivity guidelines, help ensure LLMs' adherence to task-specific requirements.

\item \textbf{Aggregation Error:} These errors occur when LLMs apply incorrect aggregation functions, often because questions contain words common to both operations and column names, like \texttt{`average'}.

\textit{Solution:} Clear column information and specific query formulations, such as stating the nature of \texttt{`allied-unrelated'} columns, guide LLMs to apply the correct aggregation method.

\item \textbf{Function-Column Ambiguity Error:} This type of error manifests when there is confusion between column names and function names, leading to incorrect query execution.

\textit{Solution:} Renaming columns may not always work. Encapsulating column names in quotes in queries can distinguish them from function commands, aiding accurate interpretation.

\item \textbf{Insufficient Column Data/Format Information Error:}
These errors often occur due to mismatches between the LLM's assumptions about dataset structure and the actual data format, particularly in handling date-related queries.

\textit{Solution:} Specifying column formats, such as the format of \texttt{`Date'} column, in prompts ensures precise LLM data handling.

\item \textbf{Coding Syntax Error:}
Highlights differences in LLMs' coding capabilities, especially in structuring and executing dataframe queries.

\textit{Solution:} Choosing an advanced base LLM or training on DataFrame QA datasets enhances their query optimization and data handling skills.

\item \textbf{Hallucination Error:}
This type of error arises when LLMs create responses based on incorrect assumptions or non-existent data, often due to a lack of domain knowledge.

\textit{Solution:} Providing detailed data and column information in prompts, like explaining how abalone age is determined, helps LLMs bridge domain knowledge gaps and improve query accuracy.

\end{itemize}
\end{itemize}

In summary, DataFrame QA tasks present inherent challenges that broadly fall into two categories. The first pertains to the inherent capabilities of LLMs, particularly visible in issues like Instruction Misalignment and Coding Syntax Errors. GPT models, in particular, have a significantly lower error rate in these areas compared to other models, showcasing their superior ability to align with human instructions and coding accuracy.

The second category comprises challenges unique to the DataFrame QA environment.  This includes complexities in question interpretation, table header naming, and domain knowledge gaps, which are central to the task-specific intricacies. Additionally, difficulties that arise from missing table value formats and value range specifications also contribute to this category, leading to errors in query processing. Tackling both the model-intrinsic limitations and these dataframe-specific complexities is vital for enhancing LLMs' performance in DataFrame QA tasks.

\section{Related Work}
Table QA is a domain of natural language processing that focuses on interpreting and answering queries based on tabular data. This field can be broadly divided into two key tasks:

\subsection{Text-to-SQL} 
This task involves converting natural language questions into SQL queries that can be executed against relational databases. The aim is to accurately interpret the user's intent and translate it into syntactically and semantically correct SQL commands. Recent advancements in Text-to-SQL have primarily leveraged neural network-based approaches, including LLMs \cite{ye2023large,ni2023lever}, especially sequence-to-sequence models \cite{liu2021tapex,xu2021sead,herzig2020tapas,yu2018syntaxsqlnet,zhong2017seq2sql}. These technologies have demonstrated significant effectiveness in understanding diverse queries across various domains and in generating the corresponding SQL statements. Current Text-to-SQL technologies primarily rely on simple database schemas and basic queries, limiting their ability to handle complex, real-world database structures and advanced relational tasks.

\subsection{QA on Semi-Structured Tables} 
This task focuses on accurately parsing HTML tables, which are often semi-structured and vary in format, to understand the context and relationships within the data and provide the correct answers \cite{pasupat2015compositional}. It requires advanced techniques in data extraction, contextual understanding, and natural language processing to effectively navigate the diverse structures and formats of HTML tables. Recent developments have utilized transformer-based models \cite{xie2022unifiedskg,pan2021cltr,glass2021capturing,yin2020tabert} and LLMs \cite{li2023table}, significantly improving the ability to process and interpret complex table structures and query contexts. These models have notably improved the accuracy and efficiency of extracting information from semi-structured HTML tables, representing a substantial advancement in the field. However, these models face limitations when loading HTML/CSV-formatted tables, often showing limited proficiency in tasks such as identifying missing cells or finding column names, leading to low accuracy in specific tests \cite{li2023table}. Including table values in input poses data privacy risks and challenges due to context window limitations and the handling of sensitive information.

\section{Conclusions}
In summary, our introduction of a new DataFrame QA task and framework represents a significant advancement in the field. This zero-shot approach, which leverages dataframe headers and datatypes along with user questions and deliberately excludes table values, addresses data privacy concerns and minimizes extraneous data in prompts. Beyond this, DataFrame QA can further enrich the prompts with dataset descriptions and column data format details, aiding in clarifying the column meanings within dataframes. This method not only offers improved control over code execution outputs but also provides greater scalability compared to traditional Text-to-SQL tasks.

Through testing with advanced open-source and closed-source LLMs, we have analyzed error patterns, challenges and determined that the efficacy of DataFrame QA relies not only on the coding abilities of LLMs but also on their understanding of the relationship between user questions, dataframe columns, and provided instructions. In particular, the accuracy rate of GPT-4 largely consistent with practical applications.

\section{Limitation and Future Work}
A current limitation of our DataFrame QA framework is its restriction to the Pandas library, lacking integration with other relevant libraries such as NumPy, scikit-learn, and Matplotlib. Moving forward, our future work aims to address this gap by expanding the variety of libraries utilized, investigating the synergistic capabilities of these libraries to solve more complex problems. Additionally, we plan to enhance our framework to support a multi-agent system, which could add significant depth and functionality. Another key area of future development involves expanding our dataset, transforming the DataFrame QA into a task that better trains LLMs in coding capabilities. This expansion will not only aim to bridge gaps in interpreting real-world questions and generating code but will also explore the inclusion of questions involving numerical calculations and statistics. 

\bibliography{anthology,custom}

\clearpage
\appendix
\appendixpage
\addappheadtotoc
\section{UCI-DataFrameQA Dataset Generation with GPT-4}
\section{Error Classification with GPT-3}
\section{Examples of Challenges in DataFrame QA and Potential Solutions}

\begin{figure*}[ht]
\centering
\includegraphics[width=\linewidth]{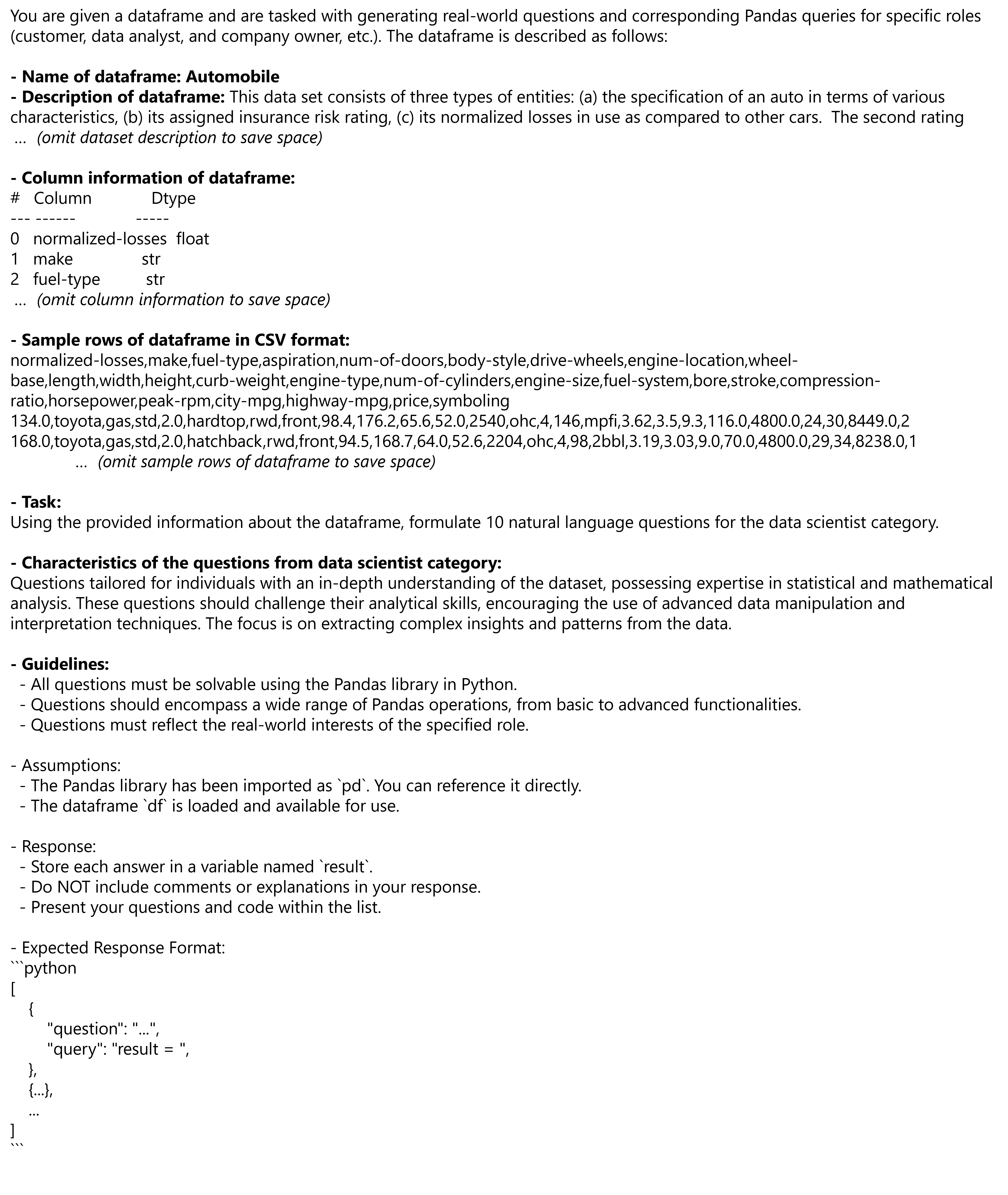}
\caption{\textbf{Sample Prompt for DataFrame QA Dataset Generation with GPT-4.}}
\label{fig:prompt_dataset_generation}
\end{figure*}

\clearpage

\begin{table*}[ht]
\small
\centering 
\caption{Characteristics of Questions from Three Different Roles Used in the Prompt.} 
\begin{tabular}{>{\raggedright\arraybackslash}p{2cm}p{12cm}}
\toprule
\textbf{Role} & \textbf{Description of Question Characteristics} \\
\midrule
Data Scientist & Questions tailored for individuals with an in-depth understanding of the dataset, possessing expertise in statistical and mathematical analysis. These questions should challenge their analytical skills, encouraging the use of advanced data manipulation and interpretation techniques. The focus is on extracting complex insights and patterns from the data. \\
\rowcolor{Gray}
General User & Questions designed for users who may not have specialized data analysis skills but are interested in the practical, consumer-oriented aspects of the data. These questions should be formulated based on the nature and context of the data, requiring inferential thinking about its potential end-users. Questions and queries should be structured to be somewhat open-ended, avoiding direct references to specific column names, thus introducing a level of interpretative ambiguity. \\
Data Owner & Questions aimed at individuals or entities who own or have created the data, with a focus on business-oriented insights. These questions should cater to their interest in understanding the broader business implications, trends, and strategic insights that can be derived from the data. The emphasis is on leveraging the data for decision-making, performance tracking, and identifying opportunities or areas for improvement within the business context. \\
\bottomrule
\label{tab:three role}
\end{tabular}
\end{table*}
\clearpage

\clearpage

\begin{table*}[ht]
\small
\centering 
\caption{Sample Generated Question/Pandas Query Pairs of UCI-DataFrameQA Dataset.} 
\begin{tabular}{>{\raggedright\arraybackslash}p{2cm}p{4cm}p{8cm}}
\toprule
\textbf{Role} & \textbf{Question} & \textbf{Pandas Query} \\
\midrule
Data Scientist & How has the average weight of cars changed over the model years? & \texttt{result = df.groupby(`model\_year')[`weight'].mean()} \\
\rowcolor{Gray}
Data Scientist & What is the distribution of tumor size for cases with recurrence events? & \begin{tabular}[t]{@{}l@{}}\texttt{result = df[df[`Class'] == `recurrence-events']} \\ \texttt{[`tumor-size'].value\_counts()}\end{tabular} \\
General User & Which cars have more than 6 cylinders? & \texttt{result = df[df[`cylinders'] > 6]} \\
\rowcolor{Gray}
General User & What is the most common tumor size observed in the data? & \texttt{result = df[`tumor-size'].mode()[0]} \\
Data Owner & What are the names of the cars with the top 3 highest fuel efficiencies in our dataset? & \texttt{result = df.nlargest(3, `mpg')[`car\_name']} \\
\rowcolor{Gray}
Data Owner & What is the frequency of tumor sizes in the age group 50-59? & \begin{tabular}[t]{@{}l@{}}\texttt{result = df[df[`age'] == `50-59'][`tumor-size']} \\ \texttt{.value\_counts()}\end{tabular} \\
\bottomrule
\label{tab:question query pairs}
\end{tabular}
\end{table*}
\clearpage

\begin{table*}[ht]
\small
\centering 
\caption{Description of Eight Pre-defined Error Classes.} 
\begin{tabular}{>{\raggedright\arraybackslash}p{2.1cm}p{5cm}p{7cm}}
\toprule
\textbf{Abbreviation} & \textbf{Error Classes} & \textbf{Description} \\
\midrule
String Error & String Matching and Comparison Errors & Errors in this class arise from improper handling of string comparisons, such as failing to use appropriate matching methods, not accounting for case sensitivity, whitespace, or special characters, and using exact matching where pattern recognition is required. \\
\rowcolor{Gray}
Access Error & Data Access and Bounds Errors & This class is for errors when data is accessed using an incorrect index or key, or when the index exceeds the bounds of the data structure, leading to \texttt{`index out of bounds'} or \texttt{`key not found'} errors. \\
Condition Error & Query Condition and Value Errors & This class covers errors where query conditions do not reflect the data accurately or the wrong values are used, resulting in no matches or incorrect results. It includes using incorrect column names or values and failing to match the query criteria with the dataset. \\
\rowcolor{Gray}
Type Error & Data Type and Operation Errors & This class includes errors from attempting operations between incompatible data types, using methods unsuitable for the data type, and applying aggregation functions incorrectly, often leading to type mismatches or operation errors on non-compatible data types. \\
Expectation Error & Expectation and Interpretation Errors & This class encompasses errors from a discrepancy between expected outcomes and actual results, which may stem from misinterpreting the output, data, or having incorrect expectations of the data's structure, leading to incorrect assumptions and results. \\
\rowcolor{Gray}
Structure Error & Data Structure Reference Errors & This class refers to errors arising from incorrect assumptions or references to the data's structure, such as referencing non-existent columns or misinterpreting the content of the data, leading to queries that do not align with the actual data format or content. \\
Function Error & Function and Method Usage Errors & Errors in this category result from misusing functions or methods outside their intended purpose, such as using a function designed for a specific operation in a context where it does not apply, or calling methods on objects they are not designed for. \\
\rowcolor{Gray}
Others & Others & The category to cover any errors that do not fit into the specific categories above, such as general mistakes in code logic or implementation that leads to unexpected results or errors. \\
\bottomrule
\label{tab:error class}
\end{tabular}
\end{table*}

\begin{figure*}[ht]
\centering
\includegraphics[width=\linewidth]{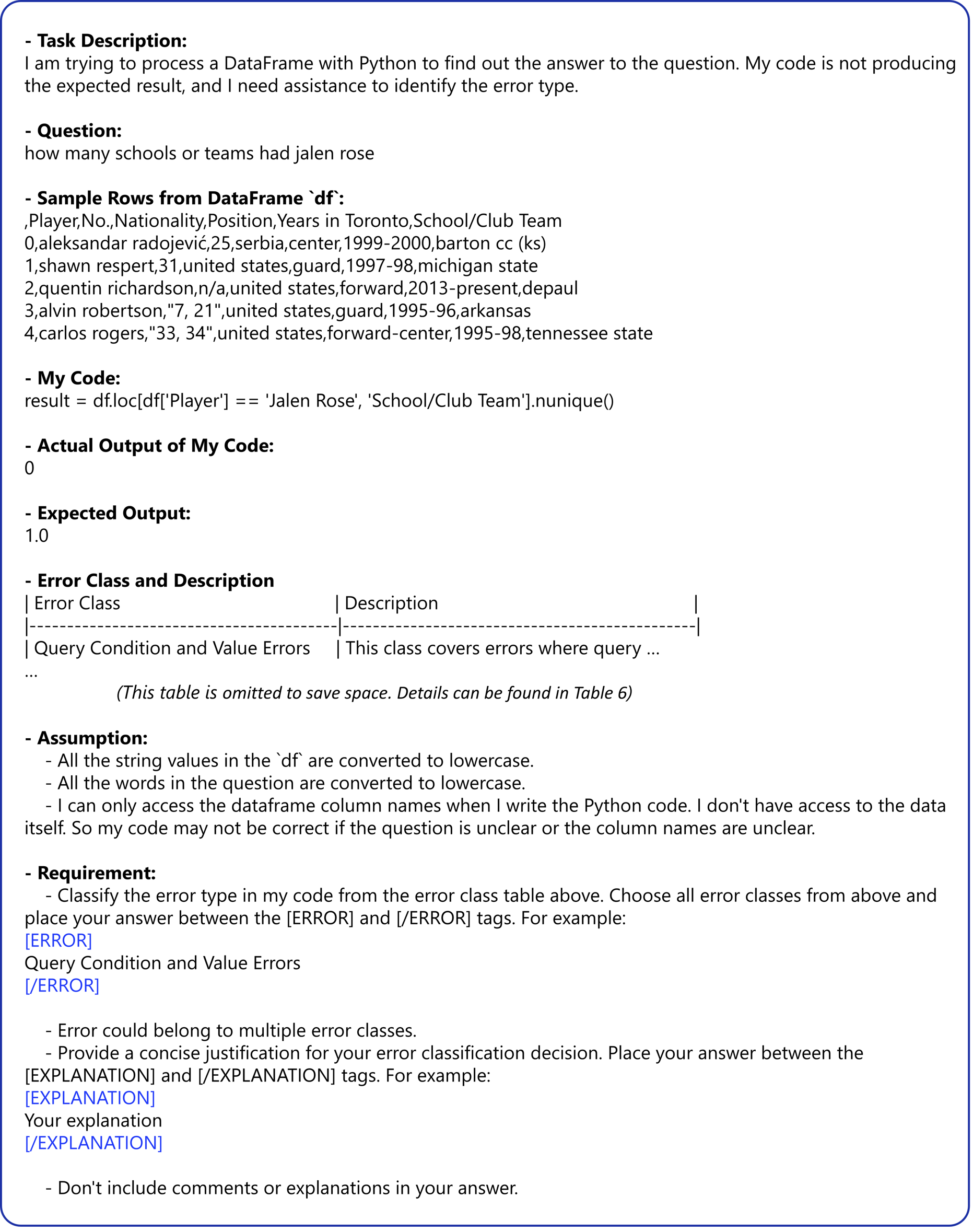}
\caption{\textbf{Sample Prompt for Error Classification with GPT-3.} Provided are the question, sample rows in the target dataframe, generated Pandas query, execution result, ground truth answer, and 8 pre-defined error classes with definitions. GPT-3 is tasked with classifying the errors present in this example. In this case, the incorrect Pandas query generated by CodeLlama-34B erroneously capitalizes the player's name when querying the dataframe.}
\label{fig:prompt_error_classification}
\end{figure*}

\begin{figure*}[ht]
\centering
\includegraphics[width=0.9\linewidth]{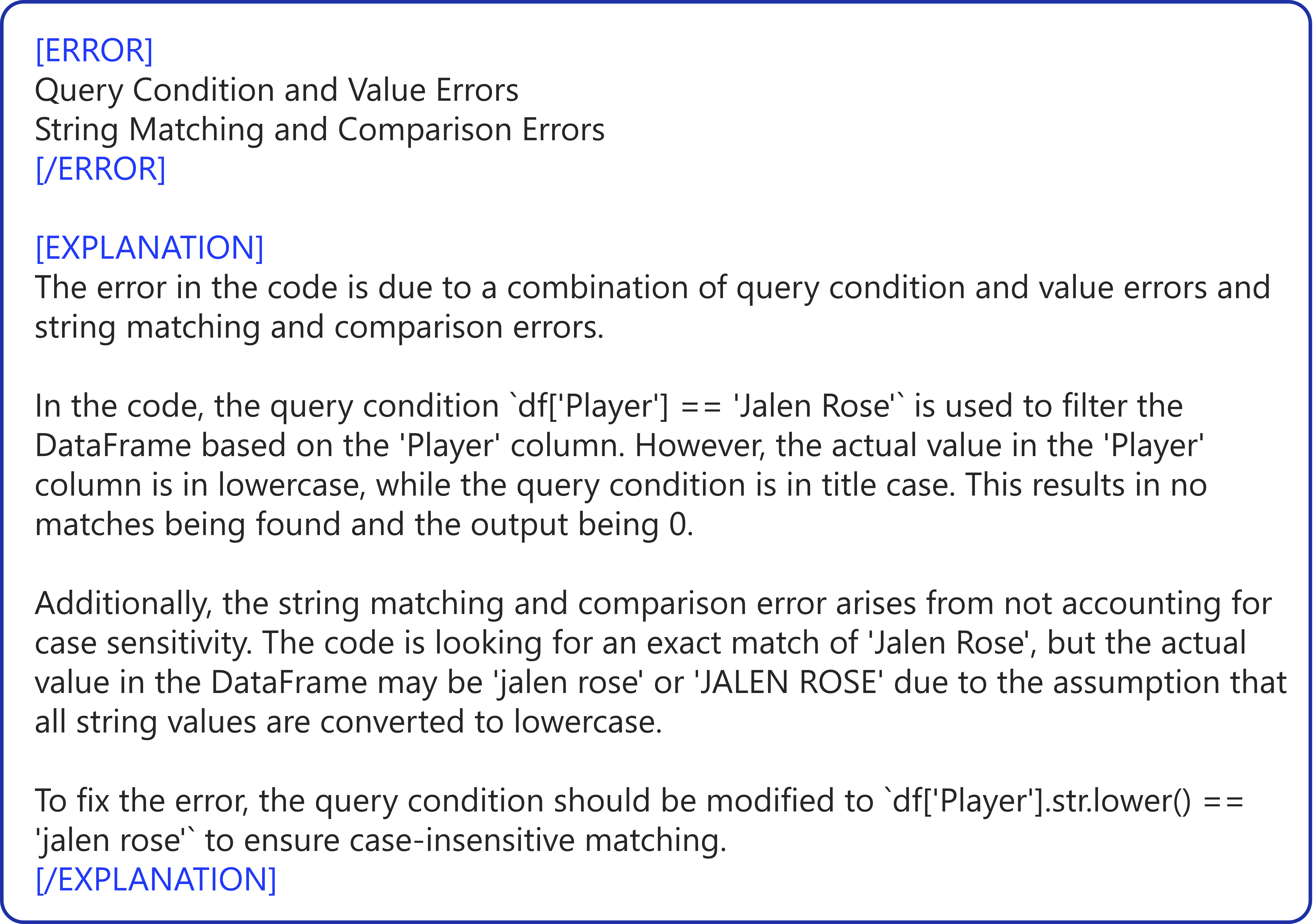}
\caption{\textbf{Sample Response for Error Classification with GPT-3.} We demonstrate the outcome of employing GPT-3.5 to assist in identifying the type of error. The response accurately categorizes the error, identifying it as a String Matching and Comparison Error, as well as a Query Condition and Value Error.\\
The analysis by GPT-3.5 underscores the importance of using lowercase search criteria for proper nouns (e.g., \texttt{`jalen rose'} instead of \texttt{`Jalen Rose'}), and recommends the implementation of the \texttt{.str.lower()} function as a standard practice. This approach ensures uniform conversion of strings to lowercase within dataframes, thereby mitigating potential case sensitivity issues. This specific case serves as a prominent example of the effectiveness of LLMs in conducting detailed error classification and analysis, demonstrating their significant value in addressing complex data processing challenges in natural language processing research.}
\label{fig:response_error_classification}
\end{figure*}

\begin{figure*}[ht]
\centering
\includegraphics[width=0.8\linewidth]{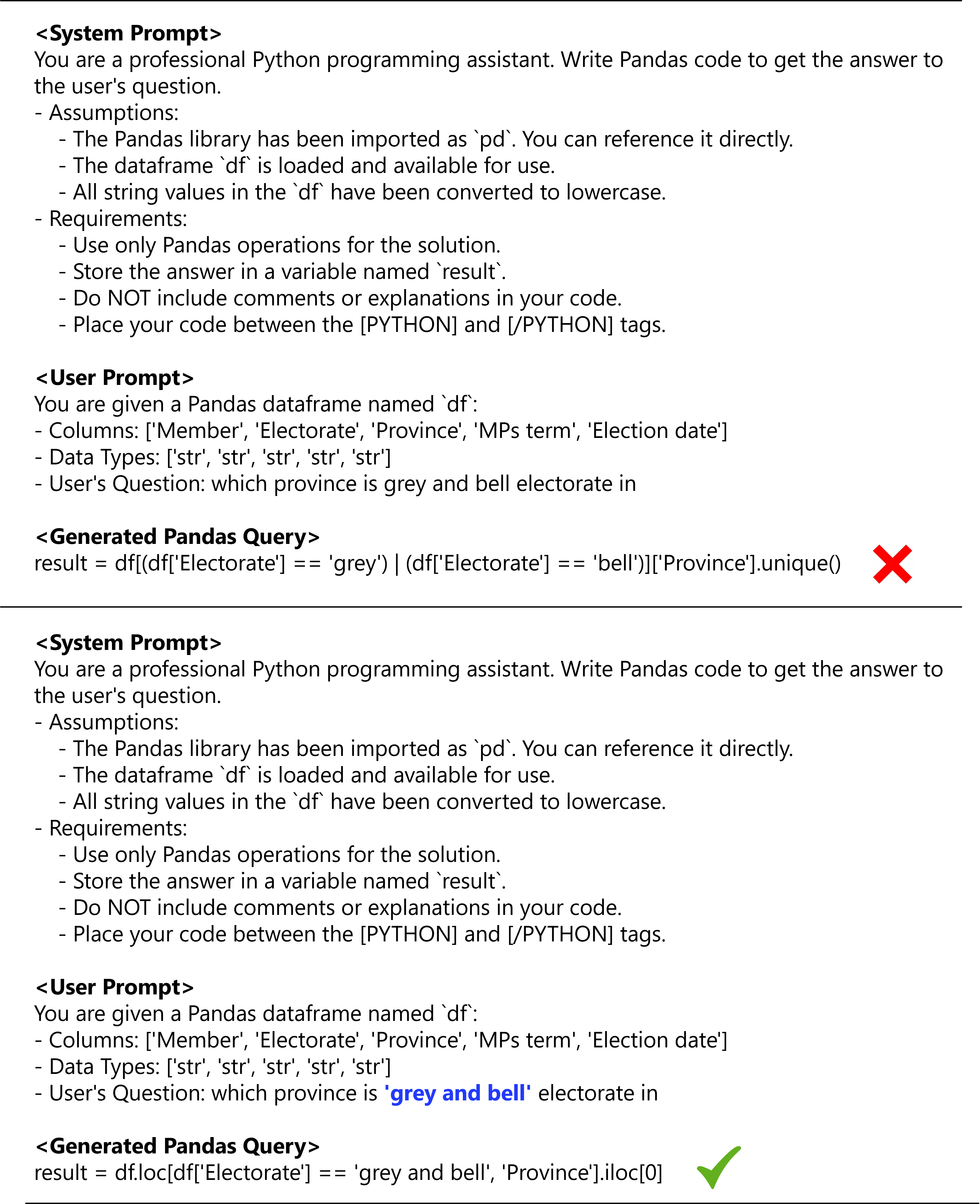}
\caption{\textbf{Value Retrieval Error by GPT-4.}
This type of error occurs in pandas queries when an incorrect value is retrieved. It often arises due to ambiguities in the user's question, leading to multiple possible interpretations. For instance, consider the example where the query targets the term \texttt{`grey and bell'} While the intended search might be for a combined entity, \texttt{`grey and bell'}, it could also be misinterpreted as two separate searches for \texttt{`grey'} and \texttt{`bell'} respectively. Another common occurrence in this error category involves GPT models inadvertently omitting special characters or symbols, such as Roman number \texttt{`I'} or hyphens, during query execution. This usually happens because the model mistakenly identifies these characters as typographical errors in the sentence, rather than integral parts of the search value. This represents one of the most common error categories in DataFrame QA tasks.\\
\textbf{Solution:} Enclosing query terms in quotation marks can significantly reduce Value Retrieval Errors. For instance, using quotations to specify \texttt{`grey and bell'} as a single entity, aiding in precise and accurate value retrieval.}
\label{fig:fc1}
\end{figure*}

\begin{figure*}[ht]
\centering
\includegraphics[width=0.8\linewidth]{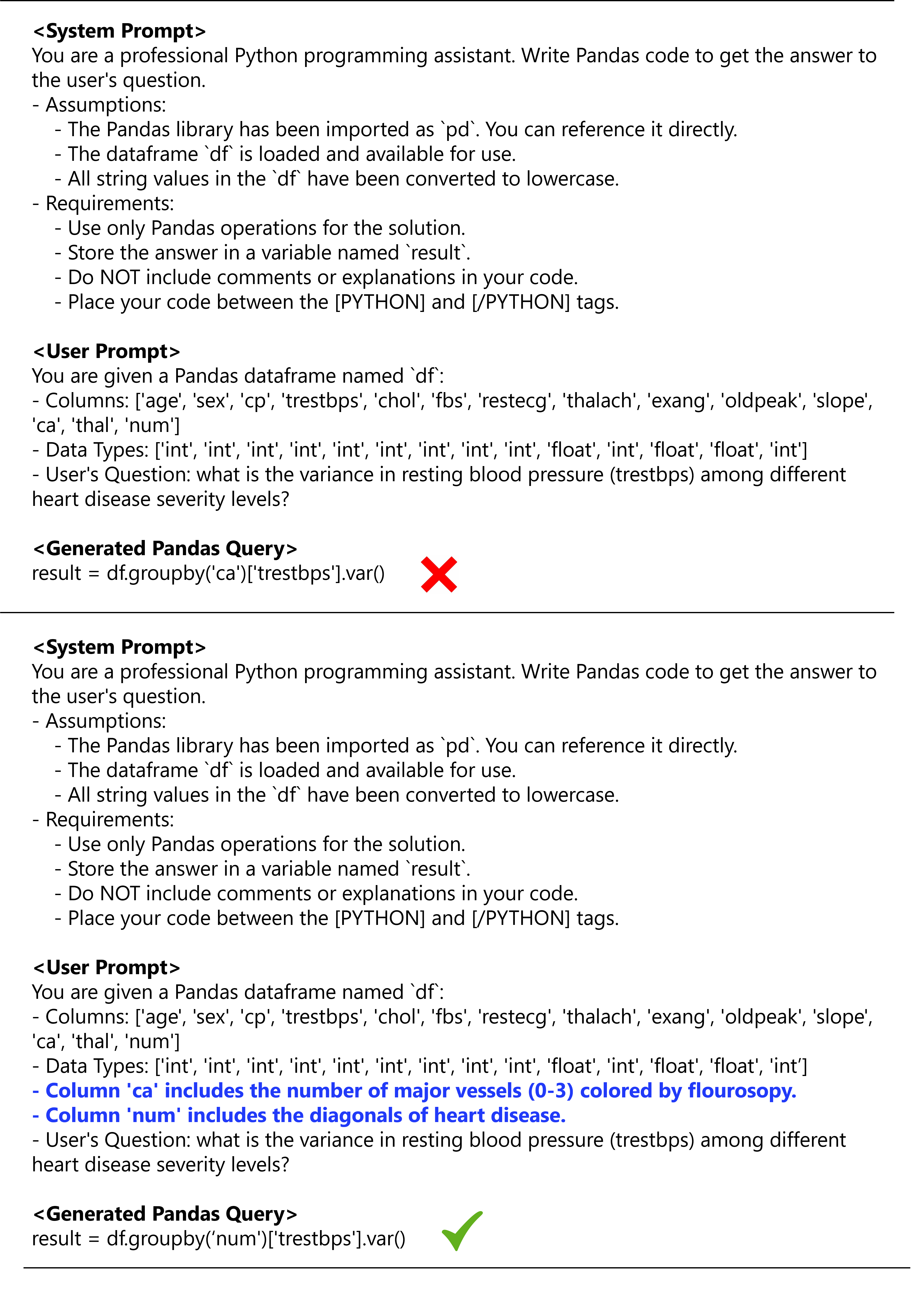}
\caption{\textbf{Column Reference Error by GPT-4}.
This error manifests when a query erroneously targets an incorrect column, commonly attributable to inadequately defined or ambiguous column names. Predominant in DataFrame QA tasks, this error underscores the imperative for a comprehensive understanding of table headers to facilitate precise data retrieval. LLMs, lacking prior domain-specific knowledge, are particularly susceptible to misidentifying columns, thereby yielding inaccurate outcomes. In the above example, the LLM fails to discern the significance of the \texttt{`num'} column, which is indicative of the diagnosis of heart disease. This oversight underscores the pivotal role of appropriate table header naming in ensuring the accuracy of DataFrame QA tasks.\\
\textbf{Solution:} Clarifying ambiguous columns in prompts can greatly reduce Column Reference Errors. For instance, stating \texttt{``Column `ca' represents major vessels colored by fluoroscopy''} and \texttt{``Column `num' indicates heart disease diagnoses''} guides LLMs to the correct data, enhancing query accuracy.}
\label{fig:fc2}
\end{figure*}

\begin{figure*}[ht]
\centering
\includegraphics[width=0.8\linewidth]{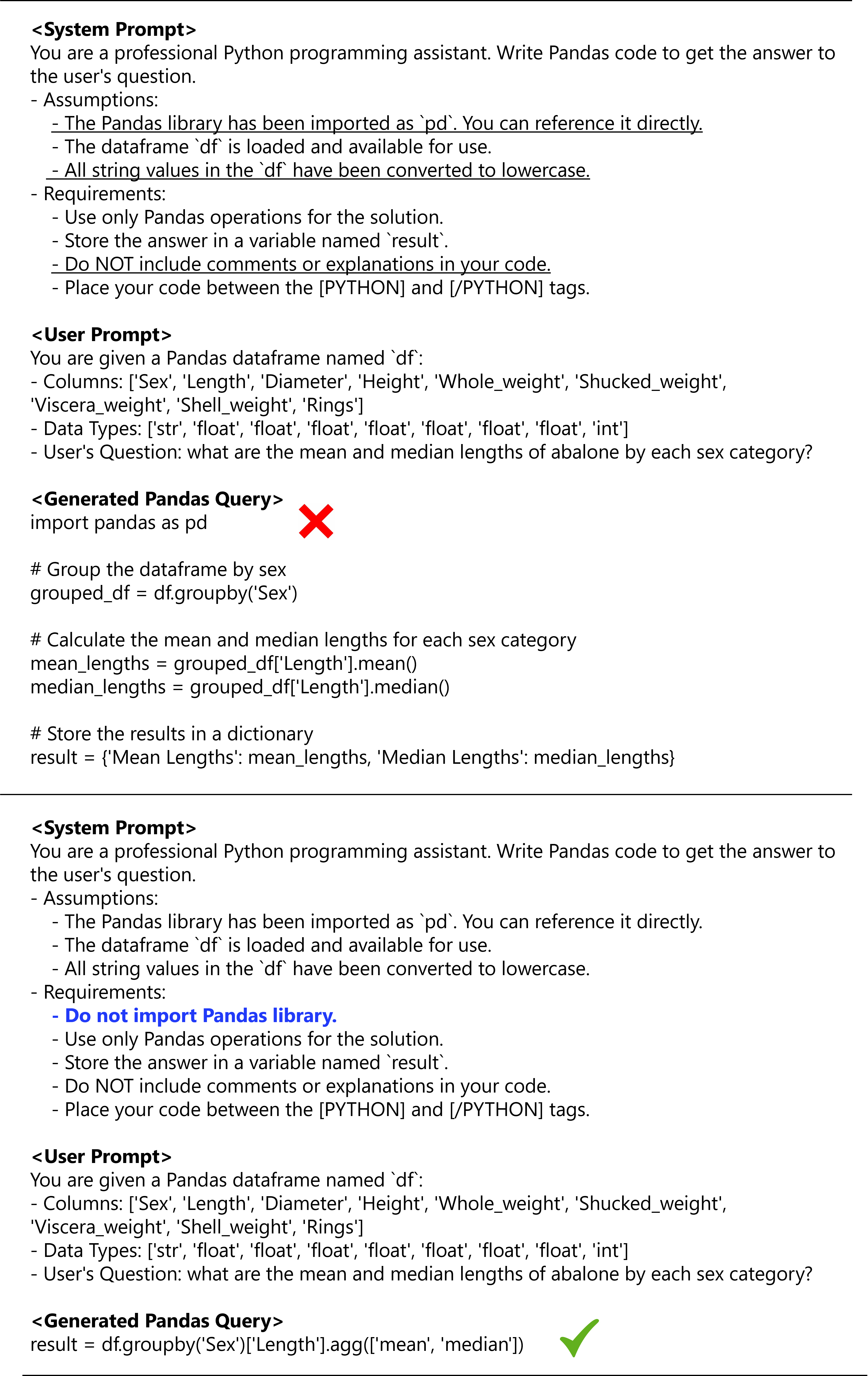}
\caption{\textbf{Instruction Misalignment Error by CodeLlama-7B.}
This error type emerges when a LLM fails to follow or comprehend given instructions. A case in the above figure, despite explicit instructions in the prompt that the pandas library can be directly utilized and that the output should not include comments, the LLM deviates from this guideline. Case sensitive cases discussed in this paper also belong to this class. This indicates a misalignment with the provided instructions, reflecting a potential bias or conflict stemming from the LLM's training on datasets where import statements and comments are standard. In our specific DataFrame QA task, such inclusions are redundant and contrary to the task requirements. This scenario exemplifies the importance of a LLM's ability to adapt to the specific nuances and requirements of a given task, distinguishing between standard programming practices and task-specific directives.\\
\textbf{Solution:} Enhancing prompts with clear directives can effectively prevent Instruction Misalignment Errors. For example, specifying \texttt{`Do not import Pandas library'} alongside \texttt{`Pandas is pre-imported as pd'} emphasizes the need for LLMs to strictly follow given instructions, improving task adherence and accuracy.}
\label{fig:fc3}
\end{figure*}

\begin{figure*}[ht]
\centering
\includegraphics[width=0.8\linewidth]{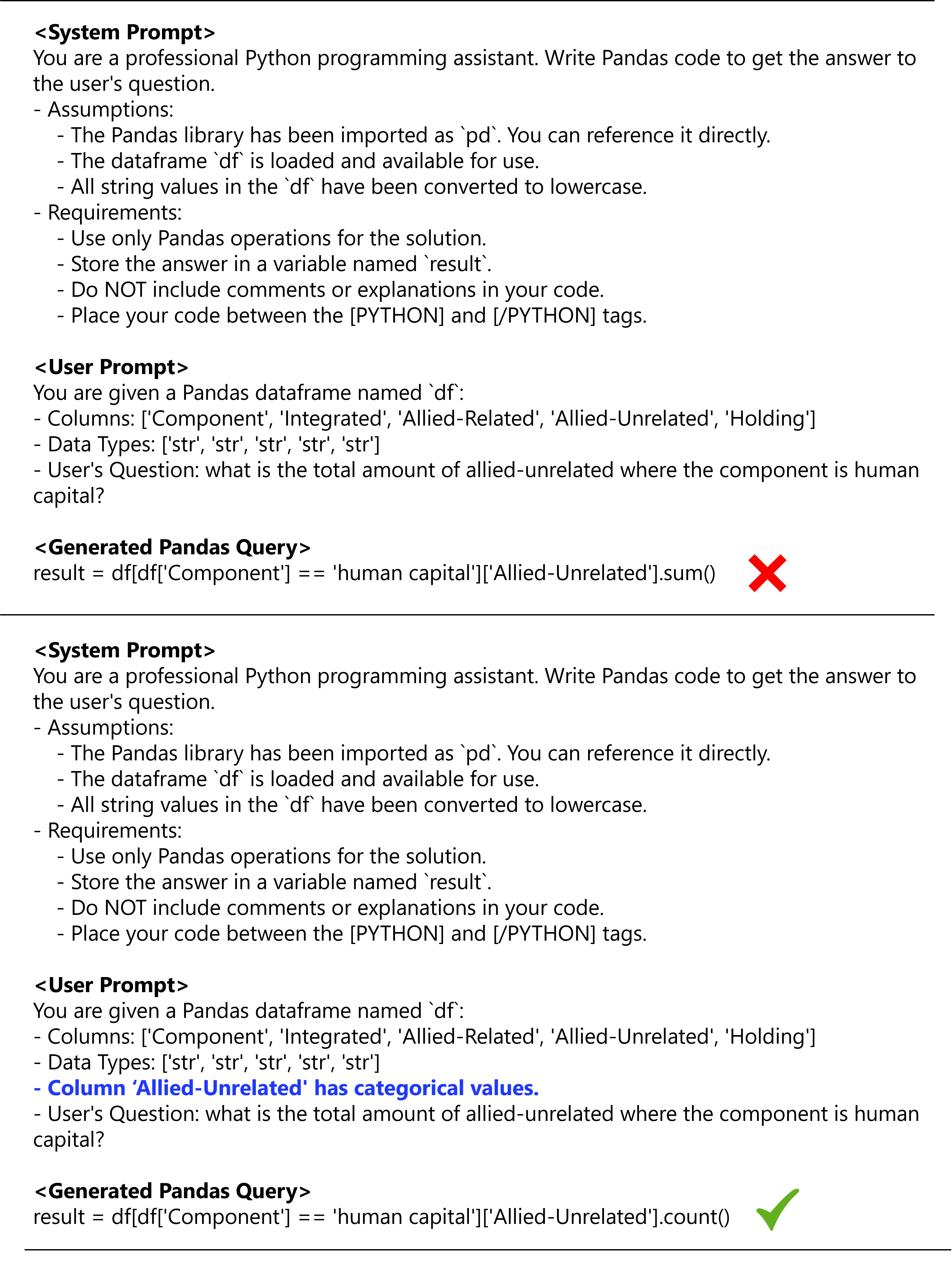}
\caption{\textbf{Aggregation Error by GPT-4.}
This error arises in dataframe queries when a LLM incorrectly applies an aggregation function due to a lack of clear understanding of the specific meaning and content of a column, compounded by the ambiguity of the question. The crux of the error lies in the LLM's inability to distinguish whether a query requires the use of a summation (sum) or a count function. For instance, in response to a query like \texttt{`What is the total amount under certain conditions?'}, it's vital to discern whether the query seeks the sum of all data meeting the criteria (using the sum() function) or merely the number of instances that satisfy the conditions (using the count() function). If a LLM does not fully grasp the nuances of the column's specific meaning and the data characteristics, or if it fails to accurately interpret the intent of the question, it may select an inappropriate aggregation function, leading to results that do not align with the actual requirements.\\
\textbf{Solution:} Providing clear column information and specific query formulations can effectively prevent Aggregation Errors. For instance, stating \texttt{``Column `allied-unrelated' holds categorical data''} or rephrasing a query to \texttt{``Count `allied-unrelated' entries for human capital''} guides the LLM to apply the correct aggregation method, enhancing result accuracy.}
\label{fig:fc4}
\end{figure*}

\begin{figure*}[ht]
\centering
\includegraphics[width=0.8\linewidth]{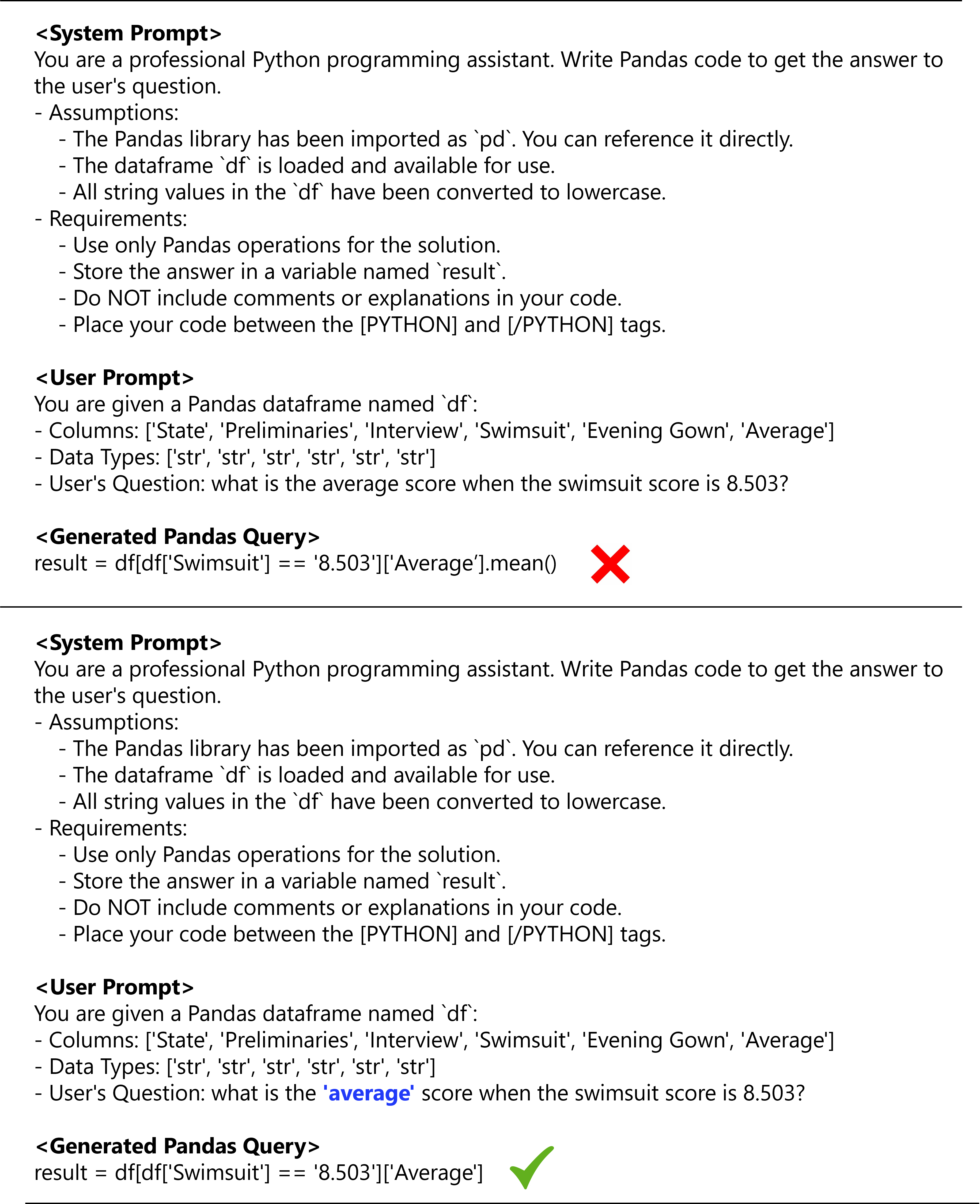}
\caption{\textbf{Function-Column Ambiguity Error by GPT-4.}
This error manifests in dataframe queries when there is ambiguity between column names and function names, leading to erroneous interpretations and executions by the LLM. A typical instance of this error is seen when a column name includes terms like \texttt{`Average'} or \texttt{`Avg'} which can also be names of aggregation functions. This leads to an unnecessary aggregation operation instead of the simple data retrieval that the query required.\\
\textbf{Solution:} Renaming columns, such as changing to \texttt{`average score'}, may not rectify Function-Column Ambiguity Errors. A preferable approach is to encapsulate column names in quotation marks within queries, clearly differentiating them from function commands, thereby guiding accurate LLM interpretation.}
\label{fig:fc5}
\end{figure*}

\begin{figure*}[ht]
\centering
\includegraphics[width=0.8\linewidth]{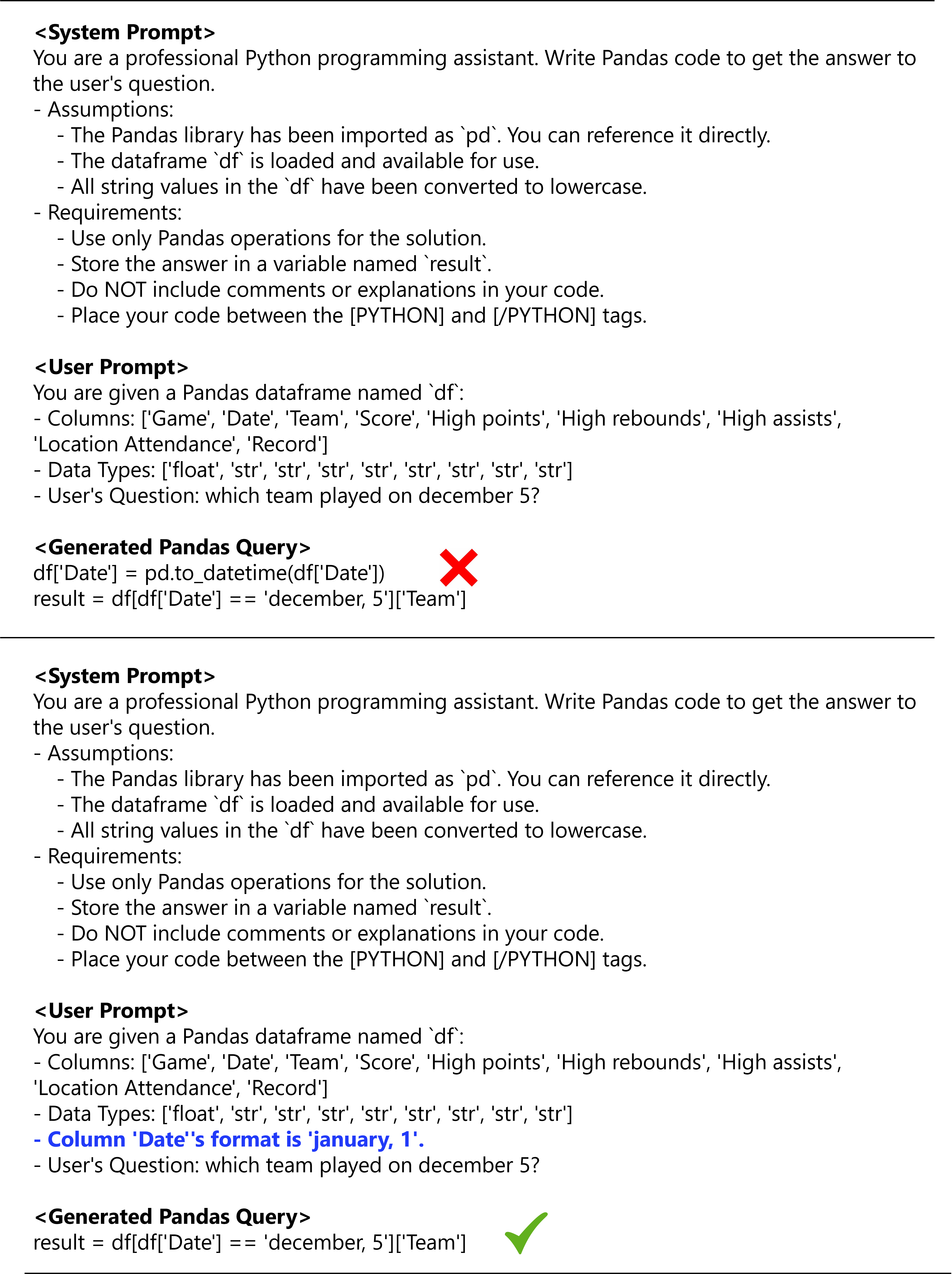}
\caption{\textbf{Insufficient Column Data/Format Information Error by GPT-4.}
This error arises when there is a mismatch between the LLM's assumptions about a dataset's structure and the actual data format, leading to incorrect dataframe operations. A notable instance of this occurs when handling date-related queries without clear information on the date format in the dataset.
LLM erroneously assumes that the \texttt{`Date'} column in the dataset contains full date information including the year. However, in reality, the dataset's \texttt{`Date'} column only contains month and day, without the year, leading to a failure in correctly applying the $to\_datetime$ function. In this case, the LLM's error stems from a gap in understanding the specific format of the \texttt{`Date'} column, highlighting the importance of having accurate and sufficient information about the data's structure and format. \\
\textbf{Solution:} Specifying column formats in prompts, such as stating \texttt{``Column `Date' follows the `january, 1' format''} effectively addresses Insufficient Column Data/Format Information Errors, ensuring precise LLM data handling.}
\label{fig:fc6}
\end{figure*}

\begin{figure*}[ht]
\centering
\includegraphics[width=0.8\linewidth]{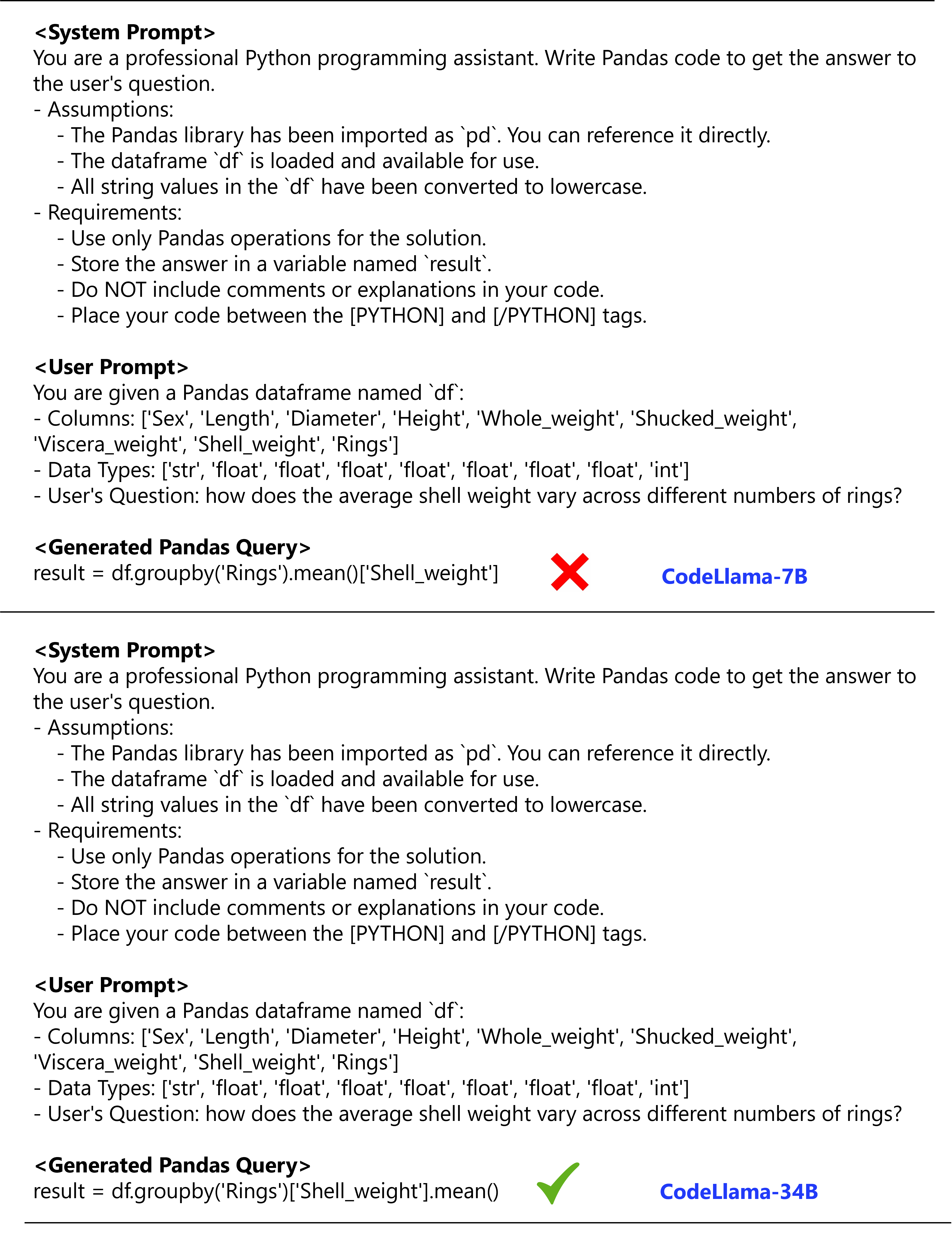}
\caption{\textbf{Coding Syntax Error by CodeLlama-7B.}
This type of error highlights disparities in the coding capabilities of LLMs, particularly in structuring and executing DataFrame queries. It occurs when the syntax used in a query is incorrect or suboptimal, impacting the query's functionality and efficiency. In the above example, the \texttt{.mean()} function is applied across all columns in the grouped DataFrame before selecting the \texttt{`Shell\_weight'} column. Such an approach is not just inefficient but also potentially problematic. If the DataFrame contains non-numerical columns, computing the mean for all columns initially can lead to errors, as the mean function is not applicable to non-numerical data. This kind of error emphasizes the challenges LLMs face in coding proficiency, particularly regarding the optimization of code for data manipulation tasks. \\
\textbf{Solution:} Addressing Coding Syntax Errors depends on the LLM's coding expertise. Solutions include choosing a base LLM with enhanced coding abilities, such CodeLlama-34B or GPT-4 or training the LLM on DataFrame QA datasets for better query optimization and data handling skills.}
\label{fig:fc7}
\end{figure*}

\begin{figure*}[ht]
\centering
\includegraphics[width=0.8\linewidth]{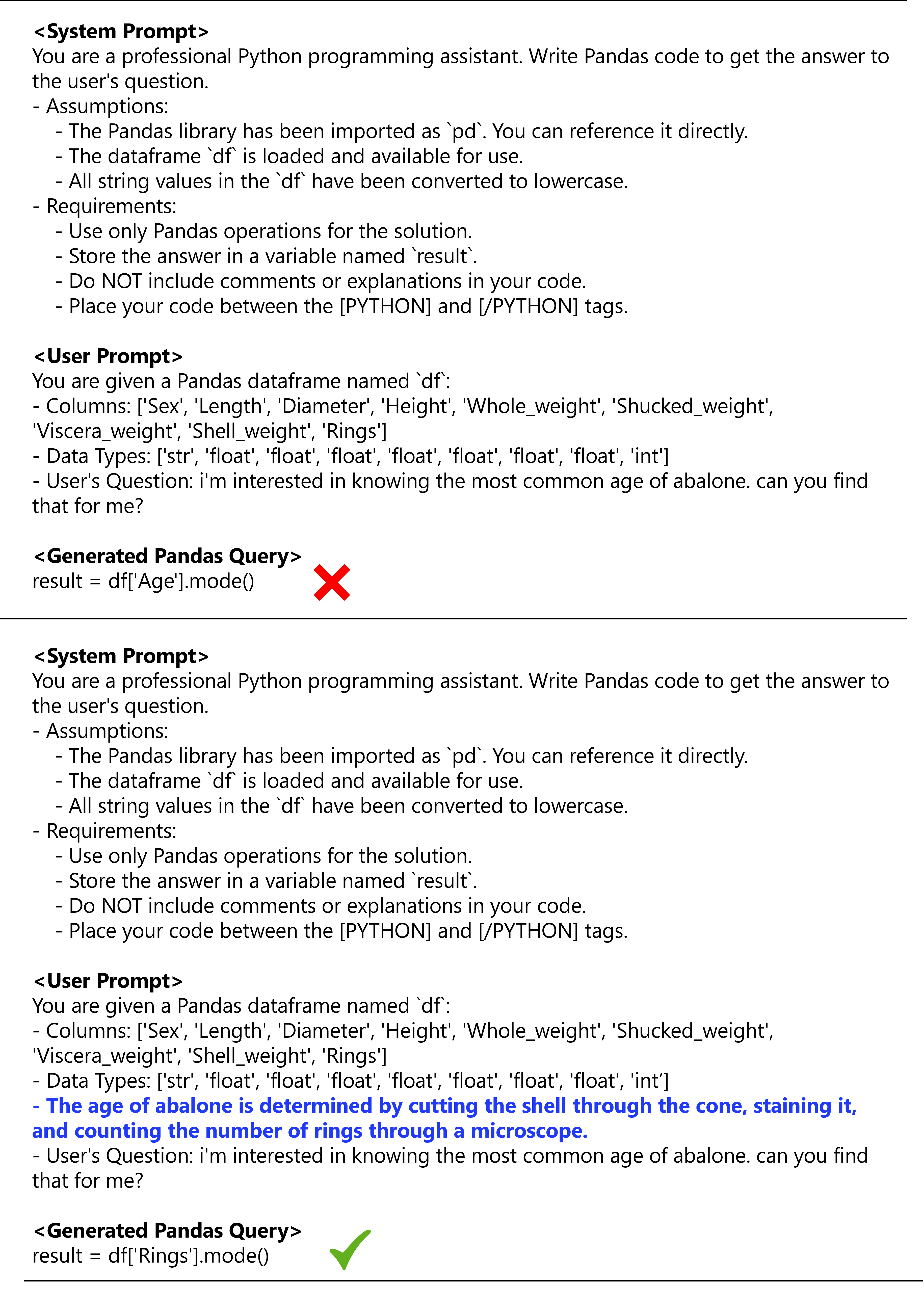}
\caption{\textbf{Hallucination Error by CodeLlama-7B.}
This error type arises when a LLM generates responses based on incorrect assumptions or fabricated details, often due to a lack of domain-specific knowledge. In DataFrame queries, this manifests as references to non-existent columns or data points that the LLM `hallucinates' or incorrectly infers. In the illstruated example, the accurate approach should involve using the \texttt{`Rings'} column, which typically represents the age of abalone. However, a Hallucination Error occurs when the LLM creates a query based on an imaginary \texttt{`Age'} column that doesn't exist in the dataset. This error is a result of the LLM's lack of understanding that in the context of abalone, age is commonly denoted by the number of rings, not a separate \texttt{`Age'} column. It demonstrates a significant gap in domain-specific knowledge, where the LLM fails to accurately interpret the data context and instead relies on incorrect or made-up information.  \\
\textbf{Solution:} Enhancing prompts with detailed data and column information, like specifying \texttt{`Abalone age is assessed by counting rings on the shell'} helps bridge domain knowledge gaps in LLMs, effectively reducing Hallucination Errors and improving query accuracy.}
\label{fig:fc8}
\end{figure*}

\end{document}